\documentclass{article}

\usepackage{arxiv}

\usepackage{cite}
\usepackage{amsmath,amssymb,amsfonts}
\usepackage{listings}
\usepackage{graphicx}
\usepackage{textcomp}
\usepackage{hyperref}
\hypersetup{pageanchor=false} 
\usepackage{svg}
\usepackage{placeins}
\usepackage{caption}
\usepackage{subcaption}
\usepackage{graphicx}
\usepackage{xspace}
\usepackage{tabularx}
\usepackage{array}
\usepackage{ragged2e}
\usepackage{todonotes}
\usepackage{comment}
\usepackage[nolist]{acronym}
\usepackage{multirow}
\usepackage{booktabs}
\usepackage{makecell}

\def\BibTeX{{\rm B\kern-.05em{\sc i\kern-.025em b}\kern-.08em
    T\kern-.1667em\lower.7ex\hbox{E}\kern-.125emX}}

\definecolor{mygreen}{rgb}{0,0.4,0}
\definecolor{mygray}{rgb}{0.5,0.5,0.5}
\definecolor{mymauve}{rgb}{0.2941,0,0.5098}
\definecolor{flame}{rgb}{0.89, 0.35, 0.13}

\newcommand{\ourmethod}{AcME-AD\xspace}
\newcommand{\newmethod}{ExIFFI\xspace}
\newcommand{\EIFplus}{ \texorpdfstring{$\text{EIF}^+$\xspace}{EIFp}}
\newcommand{\autoencoder}{AutoEncoder\xspace}

\DeclareFixedFont{\ttb}{T1}{txtt}{bx}{n}{9} 
\DeclareFixedFont{\ttm}{T1}{txtt}{m}{n}{9}  
\begin{document}

\title{
  \vspace{0.25in} Towards Transparent and Efficient Anomaly Detection in Industrial Processes through ExIFFI
  \thanks{Corresponding author: davide.frizzo.1@studenti.unipd.it}
  \thanks{This work was partially carried out within the MICS (Made in Italy - Circular and Sustainable) Extended Partnership and received funding from Next-GenerationEU (Italian PNRR - M4C2, Invest 1.3 - D.D. 1551.11-10-2022, PE00000004).}
}

\author{
  Davide Frizzo \\
  Department of Information Engineering \\
  University of Padova \\
  Padova (Italy) \\
  \And
  Francesco Borsatti \\
  Department of Information Engineering \\
  University of Padova \\
  Padova (Italy) \\
  \And
  Alessio Arcudi \\
  Department of Information Engineering \\
  University of Padova \\
  Padova (Italy) \\
  \And
  Antonio De Moliner \\
  Zoppas Industries Heating Element Technologies \\
  Vittorio Veneto (Italy) \\
  \And
  Roberto Oboe \\
  Department of Management and Engineering \\
  University of Padova \\
  Padova (Italy) \\
  \And
  Gian Antonio Susto \\
  Department of Information Engineering \\
  University of Padova \\
  Padova (Italy) \\
}

\maketitle
\begin{abstract}

\ac{AD} is crucial in industrial settings to streamline operations by detecting
underlying issues. Conventional methods merely label observations as normal or
anomalous, lacking crucial insights. In Industry 5.0, interpretable outcomes
become desirable to enable users to understand the rationale behind model
decisions. This paper presents the first industrial application of \newmethod,
a recent approach for fast, efficient explanations for the \ac{EIF} \ac{AD}
method.\ \newmethod\ is tested on four industrial datasets,
demonstrating superior explanation effectiveness, computational efficiency and
improved raw anomaly detection performances.\ \newmethod\ reaches over 90\%
of average precision on the benchmarks considered in the study and
outperforms state-of-the-art \ac{XAI} approaches in terms of the feature
selection proxy task metric which was specifically introduced to quantitatively
evaluate model explanations.

\end{abstract}


\begin{acronym}
  \acro{AI}{Artificial Intelligence}
  \acro{AD}{Anomaly Detection}
  \acro{IF}{Isolation Forest}
  \acro{EIF}{Extended Isolation Forest}
  \acro{DIFFI}{Depth Based Isolation Forest Feature Importance}
  \acro{AcME-AD}{Accelerated Model Explanations for Anomaly Detection}
  \acro{ML}{Machine Learning}
  \acro{DL}{Deep Learning}
  \acro{MLOps}{Machine Learning Operations}
  \acro{IoT}{Internet of Things}
  \acro{IIoT}{Industrial Internet of Things}
  \acro{XAI}{Explainable Artificial Intelligence}
  \acro{SHAP}{SHapley Additive exPlanations}
  \acro{GFI}{Global Feature Importance}
  \acro{LFI}{Local Feature Importance}
  \acro{PIADE}{Packaging Industry Anomaly DEtection}
  \acro{TE}{Tennessee Eastman}
  \acro{TEP}{Tennessee Eastman Process}
  \acro{SMD}{Service Machine Data}
  \acro{SCADA}{Supervisory Control and Data Acquisition}
  \acro{PLC}{Programmable Logic Controller}
  \acro{LODA}{Lightweight on-line detector of anomalies}
\end{acronym}


\section{Introduction}\label{sec:introduction}

The rapid expansion of \ac{ML} has driven advances in \ac{IoT} technologies,
enabling widespread deployment of interconnected sensor networks~\cite{gulati2022review}.
This is particularly relevant in \ac{IIoT}, which focuses on real-time monitoring of
industrial assets to optimize maintenance and production
efficiency~\cite{sisinni2018industrial}. As \ac{IoT} systems generate vast datasets,
\ac{ML} plays a key role in extracting predictive insights, reflecting the
symbiotic relationship between \ac{ML} and \ac{IIoT}~\cite{azeem2022symbiotic}.

A key application within this context is unsupervised \ac{AD}~\cite{DeMedeiros2023ASO}, especially prominent in settings where labeling data
is impractical. Among \ac{AD} methods, models based on \ac{IF}~\cite{if},
in particular \ac{EIF}~\cite{EIF} stands out due to its speed, low memory
requirements, and high performance, making it well-suited for industrial \ac{AD}~\cite{ad_survey}. However, detecting anomalies
alone is insufficient; identifying their root causes is equally important~\cite{carletti2019explainable}.
\ac{XAI} addresses this by making \ac{EIF} outputs interpretable, supporting
informed decision-making in \ac{IIoT} environments~\cite{ahmed2022artificial}.

This paper specifically evaluates the Extended Isolation Forest Feature
Importance (\newmethod) algorithm, developed by Arcudi et
al.~\cite{exiffi}, which provides a time-efficient and tailored
interpretative approach to the \ac{EIF} model. Moreover,~\cite{exiffi} introduces a modification of \ac{EIF} named \EIFplus, which improves generalization and is
also interpretable by \newmethod.

This paper extends~\cite{exiffi_rtsi} with additional and more robust
experimental results on the applicability of \ac{EIF}, \EIFplus, and
\newmethod\ in industrial environments. Specifically, the contributions are:%

\begin{itemize}%
  \item An evaluation of the explained models in terms of \ac{AD} performances, detailed in~\ref{sec:metrics}.
  \item \newmethod\ performance assessment for \ac{PIADE} is extended to all five machines in the study, as presented in~\ref{sec:GFI-PIADE}.
  \item The effect of hyperparameters such as number of trees and contamination factor on \newmethod\ is assessed via ablation studies in~\ref{sec:ablation}.
  \item The addition of two new benchmark dataset, \texttt{CoffeeData} and \ac{SMD}, introduced in~\ref{sec:coffe_data} and~\ref{sec:smd_data}.
  \item Additional experiments on two synthetic datasets to showcase key theoretical properties of multivariate feature importance of \newmethod\ in~\ref{sec:syn_data_exp}.
  \item Additional images to showcase the results of Feature Selection~(Figures \ref{sec:fs-TEP}, \ref{sec:fs-coffe}).
  \item Inference experiments on a Raspberry PI 3 Model B to assess the industrial applicability of \newmethod\ in resource constrained environments~\ref{subsec:raspberry_exp}.
\end{itemize}

The paper is organized as follows: Section~\ref{sec:related_works}
contextualizes \newmethod\ within Industry 5.0;
Section~\ref{sec:proposed_approach} describes the algorithm;
Section~\ref{sec:experimental_results} presents benchmark datasets and results.
Properties of \newmethod\ are further analyzed via synthetic datasets
(\ref{sec:syn_data_exp}) and ablation studies (\ref{sec:ablation}).
Sections~\ref{sec:limitations} and~\ref{sec:conclusions} discuss limitations,
findings, and future directions.


\section{Related Work}\label{sec:related_works}

The transition from Industry 4.0~\cite{lasi2014industry} to Industry 5.0~\cite{valette2023industry} shifts the focus from automation-driven performance to human-centric outcomes~\cite{zizic2022industry}, integrating human creativity with \ac{AI} and robotics while requiring transparency and interpretability in \ac{ML} models.

This need is addressed by implementing \ac{XAI} in industrial applications, including fault detection in machinery~\cite{brito2022explainable}, process monitoring across sectors from semiconductors to home appliances~\cite{carletti2020interpretable, feng_2020, carletti2019deep}, and predictive maintenance~\cite{simon_2021}.
\newmethod\ exemplifies this trend, offering fast and precise interpretation of \ac{EIF}-based \ac{AD} models for \ac{IIoT} environments.


Recent works on \ac{XAI} for \ac{DL} in industrial \ac{AD}, proposed transformer-based architectures for multi-sensor systems~\cite{sensor_fusion_transformer,GenAD-SM}, graph neural
networks for IT infrastructure monitoring~\cite{graph_gan}, and deep SVDD
variants~\cite{SVDD}.
Post-hoc \ac{XAI} methods like \ac{SHAP} have also been applied to \ac{DL} approaches for various domains~\cite{shap_traffic,antenna_ad}.

However, \ac{DL} methods often require large training datasets and computational resources, labeled anomalies~\cite{SVDD}, and their interpretability can be limited compared to methods specifically designed for \ac{XAI}.
This study focuses on an unsupervised and lightweight approach that is easier to integrate in resource-constrained industrial systems.
In particular, the \ac{EIF} algorithm was considered due to its excellent performance in the \ac{AD} benchmarks~\cite{ADBench}.

We evaluate \newmethod\ against various \ac{XAI} techniques, including model-specific interpretability, i.e., one that exploits the inner working of a particular algorithm, like \ac{DIFFI}~\cite{carletti2023interpretable} which is tailored for the Isolation Forest;
and model-agnostic methods that instead are independent of the underlying model~\cite{molnar2020interpretable}
such as KernelSHAP~\cite{lundberg2017unified} and \ac{AcME-AD}~\cite{acme_ad,industrialacme}.

KernelSHAP approximates \ac{SHAP} values for any model but is computationally
demanding for large datasets.\ \ac{AcME-AD} reduces this cost, making it faster
than KernelSHAP while remaining model-agnostic.

\color{black}%


\section{Proposed Approach} \label{sec:proposed_approach}

\newmethod\ leverages the structure of the \ac{EIF} forest, similarly to how
\ac{DIFFI} exploits \ac{IF}, to assess the contribution of each feature in
determining whether a sample is anomalous. Consequently,
  \newmethod\ acts as a post-hoc interpretability wrapper for \ac{EIF} and
\EIFplus, leaving the underlying model and its \ac{AD} performance unchanged.
Importance scores are computed jointly with model fitting.

Note that, as in \ac{IF}, the \ac{EIF} method hypothesizes that anomalous
samples are located in low density regions of the input space and thus are
simpler to isolate from inliers. Consequently, the algorithm focuses on
identifying isolated anomalies and is less suited to identify subtle variations
within dense clusters such as~\cite{DBSCAN}.

In \ac{EIF}, each tree $t$ of
the forest $\mathcal{T}$ consists of nodes that partition the space using a
hyperplane $\mathcal{H}_k^t$. Each hyperplane is defined by a normal vector
$\mathbf{v}_k^t$ and an intercept point $\mathbf{p}^{t}_{k}$. The
hyperplane splits the data sample $X_k^t$ into two distinct subsets
$L^{t}_{k}$ and $R^{t}_{k}$, with $L^{t}_{k} \cup R^{t}_{k} = X_k^t$ and
$L^{t}_{k} \cap R^{t}_{k} = \varnothing$. 
\newmethod\ assesses the feature importance by calculating the imbalance generated by
each node for a given sample $x$ as follows:
\begin{equation} \label{eq:imbalance_coeff}
  \mathbf{\lambda}^{t}_{k}(\mathbf{x}) =
  \begin{cases}
    \left(\frac{|X^{t}_{k}|}{|L^{t}_{k}|}\right) \text{abs}(\mathbf{v}^{t}_{k}), & \text{if } \mathbf{v}^{t}_{k} \cdot \mathbf{x} > \mathbf{v}^{t}_{k} \cdot \mathbf{p}^{t}_{k} \\
    \left(\frac{|X^{t}_{k}|}{|R^{t}_{k}|}\right) \text{abs}(\mathbf{v}^{t}_{k}), & \text{otherwise}
  \end{cases}
\end{equation}

Differently from \ac{IF}, the non axis-aligned hyperplanes of \ac{EIF} help to
better detect anomalies spread along multiple dimensions, avoiding the
\ac{IF}~\cite{ad_survey} structural biases~\cite{EIF}. Consequently
explainability is enhanced even in highly dimensional datasets with importance
scores naturally distributed among multiple important features,
as demonstrated in Section~\ref{sec:syn_data_exp}.

The feature importance for a sample $x$ within a specific tree $t$ is
determined by summing the importance vectors from all nodes $k$ that
$x$ traverses on its path to the leaf, and the overall importance
across the entire forest is obtained by aggregating over all trees:
\begin{align}
  \mathbf{I}_{t}(x) = \sum_{k \in \mathcal{P}^{t}_x} \mathbf{\lambda}_{k}^{t}(x), \qquad
  \mathbf{I}(x) = \sum_{t \in \mathcal{T}} \mathbf{I}_{t}(x) \label{eq:importance_overall}
\end{align}

$\mathbf{I}(x)$ quantifies each feature's contribution to the isolation of $x$.
To correct for biases caused by features being sampled more frequently,
$\mathbf{I}(x)$ is normalized by the sum of vectors orthogonal to the
hyperplanes of the nodes that $x$ traverses in each tree:
\begin{align}
  \mathbf{V}(x) = \sum_{t \in \mathcal{T}} \sum_{k \in \mathcal{P}^{t}} \mathbf{v}_k^t, \qquad
  \mathbf{LFI}(x) = \frac{\mathbf{I}(x)}{\mathbf{V}(x)} \label{eq:lfi_eq}
\end{align}

The resulting \ac{LFI} score identifies which features most influence the
classification of $x$ as anomalous, enabling targeted root cause analysis.
Grouping together the importance vectors assigned to all samples,
\newmethod\ also provides a \ac{GFI}, a single vector quantifying the
overall importance of each feature in discriminating between inliers and outliers:
\begin{equation} \label{eq:gfi_eq}
  \mathbf{GFI}=\frac{\hat{\mathbf{I}}_O}{\hat{\mathbf{I}}_I}
\end{equation}

where $\hat{\mathbf{I}}_O$ and $\hat{\mathbf{I}}_I$ are the importance vectors
computed over the set of outliers and inliers respectively.


For what concerns the practical applications of \newmethod\, it can be
integrated into a \ac{MLOps} pipeline, enabling automatic visualization of
fault root causes, accelerating response to potentially disruptive events.


Domain experts can validate explanations or correct misidentified root causes,
which is especially important in unlabeled settings such as \ac{PIADE}
(Section~\ref{sec:datasets}), as detailed in~\cite{exiffi_rtsi} were \ac{GFI}
ranking results were validated by the opinion of experienced machine's
operators.


\section{Experimental Results}\label{sec:experimental_results}


This section presents the results of applying \newmethod\ to
three publicly accessible and a private dataset derived from
industrial processes, which serve as benchmarks for evaluating \newmethod's
effectiveness within real-world contexts. The datasets include \ac{TEP}, which
offers synthetic data with established ground truth for anomaly-inducing
features~\cite{tep_dataset}, \ac{PIADE}, which encapsulates typical real-world
challenges such as unlabeled and high-dimensional
data~\cite{diego2022packaging}.\
\ac{SMD}~\cite{OmniAnomaly},
instead, contains service logs recorded by a large internet company and is
equipped with anomaly labels and ground truth knowledge on anomalies' root
causes. Finally, \texttt{CoffeeData} is a confidential dataset containing
sensor measurements recorded during the brewing phase of capsules on coffee
machines.

The section is organized as follows: firstly, datasets are presented
in~\ref{sec:datasets}, the experimental setup is provided
in~\ref{sec:experimental_setup}, \ac{AD} model performances are reported
in~\ref{sec:metrics}, global interpretability is assessed
in~\ref{sec:GFI-TEP},\ \ref{sec:GFI-coffe} respectively, and four different
interpretability models are compared in~\ref{sec:fs-TEP},\ \ref{sec:fs-coffe}
by means of the Feature Selection Proxy Task.

The outcomes here discussed are achieved using Python as the base language to
implement the method and C to optimize functions embedded within the
performance-critical segments of the Python code.\footnote{The source code of
this project is available in a public repository, with reproducible results:
\url{https://github.com/AMCO-UniPD/exiffi-ind.git}.}

\subsection{Industrial IoT datasets}\label{sec:datasets}

Below we describe the four benchmark datasets (\ac{TEP}, \ac{PIADE},
\ac{SMD}, and \texttt{CoffeeData}) and their key characteristics.

\ac{PIADE} consists of machine alarms and statuses, while \ac{TEP},
\ac{SMD}, and \texttt{CoffeeData} are time-series datasets.
\ac{PIADE}, \ac{SMD}, and \texttt{CoffeeData} are collected
from real machines, while \ac{TEP} is simulation-based.


\begin{table}[ht]
  \centering
  \caption{The main characteristics of the industrial datasets used for this study are reported: dataset size, sampling rate, source and presence of labels}
  \label{tab:datasets_tab}
  \resizebox{\columnwidth}{!}{
    \begin{tabular}{cccccc}
      \toprule
      \textbf{Dataset} & \textbf{Samples} & \textbf{Features} &%
      \begin{tabular}{c}
           \textbf{Sampling} \\
           \textbf{Period} $[s]$
      \end{tabular} &%
      \begin{tabular}{c}
           \textbf{Anomaly} \\
           \textbf{Labels}
      \end{tabular} &%
      \text{Source} \\
      \midrule
      \ac{PIADE} & 23376 & 165 & 3600 & No & \cite{diego2022packaging} \\
      \ac{TEP} & 35600 & 52 & 180 & Yes & \cite{tep_dataset} \\
      \ac{SMD} & 56956 & 38 & 60 & Yes & \cite{OmniAnomaly} \\
      \texttt{CoffeeData D1} & 372 & 15 & 0.5 & Yes & confidential \\
      \texttt{CoffeeData D2} & 255 & 15 & 0.5 & Yes & confidential \\
      \bottomrule
    \end{tabular}
  }
\end{table}



A summary of the main datasets characteristics is reported in
Table~\ref{tab:datasets_tab}.


We analyze pairwise dependencies among sensor features using both Pearson
correlation and mutual information. For example in \ac{TEP} approximately 41\%
of feature pairs exhibit negligible linear correlation ($|\rho| < 0.05$), a
large majority (89\%) shows non-zero mutual information. Notably, 31\% of
feature pairs combine near-zero correlation with positive mutual information,
indicating substantial nonlinear dependencies among sensor measurements,
a challenging aspect for root cause analysis.

For a more detailed description of the \ac{TEP} and \ac{PIADE} datasets please
refer to~\cite{exiffi_rtsi}.

\subsubsection{SMD Dataset}\label{sec:smd_data}

The \ac{SMD} dataset is a publicly available multivariate time series dataset
introduced in~\cite{OmniAnomaly}. The data were collected over a period of 5
weeks by a large Internet company, encompassing 28 distinct service machines
organized into three entity groups. Each machine's data is split into two halves:
the first is used for training and the second for testing. Following the original
dataset configuration, each machine constitutes a separate and independent
dataset. For this study, we focus exclusively on the \texttt{machine-1-1} subset.

Anomalies in this dataset occur as contiguous time intervals rather than
isolated points, posing a challenge for isolation-based methods
(Section~\ref{sec:limitations}).

\subsubsection{\texttt{CoffeeData} Dataset}\label{sec:coffe_data}

The \texttt{CoffeeData} dataset was firstly introduced in~\cite{coffe_paper}.
It is a confidential dataset and consists of multiple time-series representing
the brewing process of capsules in coffee machines. The goal is to detect
anomalous capsule usages early to improve brewing quality. Anomalous behaviors
include using the same capsule twice or more times (i.e. \texttt{reused
capsule}) or using the coffee machine without any capsule inside it (i.e.
\texttt{no capsule}).


Following the procedure introduced in~\cite{coffe_paper}, only the water flow
rate signal was considered, as it was the most informative one.

Each time series is trimmed to the first 15 samples of the pre-infusion phase,
as this segment provides enough information to address the problem effectively
while also allowing for early detection of anomalies, which is critical for
practical use.

Rather than hand-crafted features as in~\cite{coffe_paper}, raw time samples
are used directly, with timestamps as features ranked by \newmethod.
This allows identifying the time instants at which anomalies occur, enabling
early alarm triggering for improper equipment usage.

Table~\ref{tab:coffe_dataset} outlines the dataset composition:
\texttt{standard} capsules are part of the inliers group (comprising both
original and compatible brands), while there are two kinds of anomalous
samples: \texttt{reused capsules} and \texttt{no capsules}.




\begin{table}[ht]
\centering
\caption{\texttt{CoffeeData} dataset composition.}
\label{tab:coffe_dataset}
\begin{tabular}{cccc}
\toprule
\textbf{Brewing} & \textbf{Capsule} & \textbf{\# Samples} & \textbf{Class} \\
\midrule
\multirow{1}{*}{1st}
  & \texttt{standard}   & 71  & \multirow{2}{*}{Normal} \\
\multirow{1}{*}{2nd}
  & \texttt{standard}   & 229 &                         \\
\midrule
\multirow{2}{*}{any}
  & \texttt{reused capsule} & 78 & \multirow{2}{*}{Anomaly} \\
  & \texttt{no capsule}     & 26 &                          \\
\bottomrule
\end{tabular}%
\end{table}

Normal capsules are divided into the first and second brewings categories
as defined in~\cite{coffe_paper}.


Second-brewing trials require no cool-down and are thus more frequent,
resulting in a higher sample count.

\subsection{Experimental Setup}\label{sec:experimental_setup}

To assess \newmethod\ both qualitatively and quantitatively, three types of
experiments are conducted:%

\begin{itemize}%
  \item \textbf{Global Importance Assessment}: This experiment produces a Score Plot ranking the features in decreasing
    order of \ac{GFI} score. It represents a powerful tool for human operators to discover the root causes of anomalies and
    take preventive actions.
  \item \textbf{Local Scoremaps}: Local scoremaps focus on local interpretability by selecting a pair of features and showing
    how inliers and outliers are distributed along them. An heatmap is superimposed to showcase the distribution of the \ac{LFI}
    scores in the feature space.
  \item \textbf{Feature Selection Proxy Task}: In this study feature selection is used as a proxy task to provide a quantitative
    measure of the effectiveness of a \ac{XAI} approach by plotting the trend of the average precision metric as features are progressively
    removed in increasing or decreasing order of \ac{GFI} scores.
\end{itemize}

To account for stochasticity, all experiments are repeated with multiple random
seeds to ensure statistical robustness.

For the sake of space results on the Local Scoremaps experiments are not
included in this extension. The interested reader can find a detailed analysis
of local interpretability in~\cite{exiffi_rtsi}.


\subsection{\texorpdfstring{\ac{AD}}{AD} Performance Comparison} \label{sec:metrics}

In this section \ac{IF}, \ac{EIF} and \EIFplus\ \ac{AD} models are compared with
typical \ac{AD} metrics for the labeled datasets (i.e.\ \ac{TEP} and
\texttt{CoffeeData}). Tables~\ref{tab:metrics_table_tep},
\ref{tab:metrics_table_coffe} report the test set metrics. This evaluation is a
crucial step for the application of \ac{XAI} methods like \newmethod\ since
their effectiveness highly relies on the accuracy of the explained model.



\newcolumntype{A}{>{\centering}p{0.09\textwidth}}
\newcolumntype{C}{>{\centering\arraybackslash}p{0.07\textwidth}}

\begin{table}[ht]
\centering
\caption{\ac{TEP} dataset \ac{AD} metrics for the considered models. Best performance for each metric in bold.}
\label{tab:metrics_table_tep}
\begin{tabular}{cccccc}
\toprule
\textbf{Model} &
\makecell{\textbf{Average} \\ \textbf{Precision}} &
\textbf{Precision} &
\makecell{\textbf{ROC} \\ \textbf{AUC}} &
\makecell{\textbf{Fit} \\ \textbf{Time [s]}} &
\makecell{\textbf{Predict} \\ \textbf{Time [s]}} \\
\midrule
\texttt{IF} & 0.85 & 0.90 & 0.92 & 1.04 & 6.00e-5 \\
\texttt{EIF} & 0.93 & 0.90 & \textbf{0.95} & \textbf{0.93} & 7.00e-5 \\
\texttt{\EIFplus} & 0.93 & \textbf{0.91} & \textbf{0.95} & 1.22 & 6.00e-5 \\
\autoencoder & 0.92 & \textbf{0.94} & \textbf{0.95} & 8.83 & 2.00e-5 \\
\texttt{DeepSVDD} & 0.79 & 0.87 & 0.89 & 1.66 & \textbf{1.00e-6} \\
\bottomrule
\end{tabular}%
\end{table}



Table~\ref{tab:metrics_table_tep} shows how \ac{EIF} and \EIFplus\ perform
better than \ac{IF} thanks to their higher-complexity data partitioning scheme,
leveraging multidimensional splits as opposed to \ac{IF}.


The fit and predict times of \ac{EIF} and \EIFplus\ are higher than \ac{IF}
because they simultaneously accumulate importance scores during training. However,
this design choice yields a significant benefit at inference time: the
importance scores are directly available without requiring an additional
explanation phase. As reported in Table~\ref{tab:tab_time_comparison}, this
results in substantially lower explanation times for the proposed approach
compared to model-agnostic methods.

Two \ac{DL}-based \ac{AD} models are also considered for comparison: \autoencoder\ \cite{AE}
and \texttt{DeepSVDD}~\cite{DeepSVDD}. While the \autoencoder\ achieves slightly higher average
precision and ROC AUC scores than \EIFplus, it requires substantially longer
training times, rendering it impractical for time-sensitive industrial
applications.

\texttt{CoffeeData} comprises two evaluation datasets: \texttt{D1}, which
contains both first and second brewings and both kinds of anomalies (i.e.
\texttt{no capsule} and \texttt{reused capsule}); while \texttt{D2} is more
limited, but equivalent to the one used in previous work~\cite{coffe_paper}
(i.e.\ only second brewing data and \texttt{no capsule} anomalies).



\begin{table}[ht]
\centering
\caption{\texttt{CoffeeData} dataset \ac{AD} metrics for the considered models. Best performance for each metric in bold.}
\label{tab:metrics_table_coffe}
\resizebox{\columnwidth}{!}{%
\begin{tabular}{ccccccc}
\toprule
\textbf{Dataset} &
\textbf{Model} &
\makecell{\textbf{Average} \\ \textbf{Precision}} &
\textbf{Precision} &
\makecell{\textbf{ROC} \\ \textbf{AUC}} &
\makecell{\textbf{Fit} \\ \textbf{Time [s]}} &
\makecell{\textbf{Predict} \\ \textbf{Time [s]}} \\
\midrule
\multirow{3}{*}{\textbf{D1}}
  & \texttt{IF}        & 0.74 & 0.65 & 0.63 & 1.18 & 5.48e-5 \\
  & \texttt{EIF}       & 0.77 & \textbf{0.66} & 0.63 & 0.81 & 3.65e-5 \\
  & \texttt{\EIFplus}  & \textbf{0.79} & 0.65 & \textbf{0.64} & 0.99 & 4.87e-5 \\
  & \texttt{AutoEncoder} & 0.44 & 0.43 & 0.62 & 0.36 & 3.0e-05 \\
  & \texttt{DeepSVDD} & 0.34 & 0.33 & 0.55 & \textbf{0.13} & \textbf{3.0e-05} \\
\midrule
\multirow{3}{*}{\textbf{D2}}
  & \texttt{IF}        & 0.91 & \textbf{0.96} & 0.97 & 0.96 & 3.04e-5 \\
  & \texttt{EIF}       & 0.97 & 0.89 & 0.98 & 0.74 & 5.48e-5 \\
  & \texttt{\EIFplus}  & \textbf{0.99} & \textbf{0.96} & \textbf{0.99} & 0.96 & 4.26e-5 \\
  & \texttt{AutoEncoder} & 0.76 & 0.36 & 0.87 & 0.32 & 4e-05 \\
  & \texttt{DeepSVDD} & 0.84 & 0.72 & 0.91 & \textbf{0.23} & \textbf{3e-05} \\
\bottomrule
\end{tabular}
}
\end{table}

\ac{AD} model performances are showcased in
Table~\ref{tab:metrics_table_coffe}. The increased difficulty associated with
\texttt{D1} translates in slightly worse \ac{AD} metrics, but still the best
performing model is \EIFplus\ followed by \ac{EIF} and \ac{IF}.
\texttt{D2} metrics match those in~\cite{coffe_paper}, despite using raw time
samples instead of hand-crafted features.

It is worth noting that the two \ac{DL}-based models achieve significantly
lower performance than the isolation-based detectors. In particular,
\texttt{AutoEncoder} and \texttt{SVDD} struggle on small datasets such as
\texttt{CoffeeData}, likely due to their large model capacity and corresponding
data requirements. This confirms that these models are not well-suited for
industrial settings with limited data availability.

The shorter execution times of \texttt{AutoEncoder} and \texttt{DeepSVDD} on
\texttt{CoffeeData} are due to its limited size (comparable to the bootstrap
size $\psi=256$ used in ensemble models). However, as shown in
Tables~\ref{tab:metrics_table_tep} and~\ref{tab:metrics_table_smd}, deep
learning-based models become significantly less efficient as data size
increases.


\newcolumntype{A}{>{\centering}p{0.09\textwidth}}
\newcolumntype{C}{>{\centering\arraybackslash}p{0.07\textwidth}}

\begin{table}[ht]
\centering
\caption{\ac{SMD} dataset \ac{AD} metrics for the considered models. Best performance for each metric in bold.}
\label{tab:metrics_table_smd}
\begin{tabular}{cccccc}
\toprule
\textbf{Model} &
\makecell{\textbf{Average} \\ \textbf{Precision}} &
\textbf{Precision} &
\makecell{\textbf{ROC} \\ \textbf{AUC}} &
\makecell{\textbf{Fit} \\ \textbf{Time [s]}} &
\makecell{\textbf{Predict} \\ \textbf{Time [s]}} \\
\midrule
\texttt{IF} & 0.36 & 0.39 & 0.64 & 0.79 & 5.00e-5 \\
\texttt{EIF} & 0.54 & 0.61 & 0.74 & \textbf{0.65} & 6.00e-5 \\
\texttt{\EIFplus} & 0.55 & 0.62 & 0.75 & 1.02 & 5.00e-5 \\
\autoencoder & \textbf{0.64} & \textbf{0.63} & \textbf{0.80} & 8.03 & 2.00e-5 \\
\texttt{DeepSVDD} & 0.52 & 0.60 & 0.73 & 1.84 & \textbf{1.00e-6} \\
\bottomrule
\end{tabular}%
\end{table}

Given the temporal clustering of anomalies characteristic of \ac{SMD}
(Section~\ref{sec:smd_data}) and the dataset's substantial size, \ac{DL}-based
detectors achieve superior detection performance. Nevertheless, this comes at
the cost of significantly higher computational requirements. During inference,
however, their execution time is lower with the respect to ensemble methods, as
they do not need to aggregate predictions from multiple ensemble components.


Overall, \ac{EIF} and \EIFplus\ outperform \ac{IF} by several percentage points
across all \ac{AD} metrics. The ensemble-based models considered in the benchmark
provide excellent detection performances and are characterized by fast
computational times, making them suitable for industrial applications as
showcased in Section~\ref{subsec:raspberry_exp}. On the other hand,
the \ac{DL}-based \autoencoder\ and \texttt{DeepSVDD} models, although they
overperform ensemble-based models on large-scale datasets, are highly
inefficient in terms of both fitting and importance computation times and thus
impractical for industrial applications.

\subsection{Case Study I:\ \texorpdfstring{\ac{TEP}}{TEP} Dataset} \label{sec:results-TEP}

\subsubsection{Global Feature Importance}\label{sec:GFI-TEP}

\begin{figure}[htbp]
    \centering
    \centerline{\includegraphics[width=0.97\linewidth]{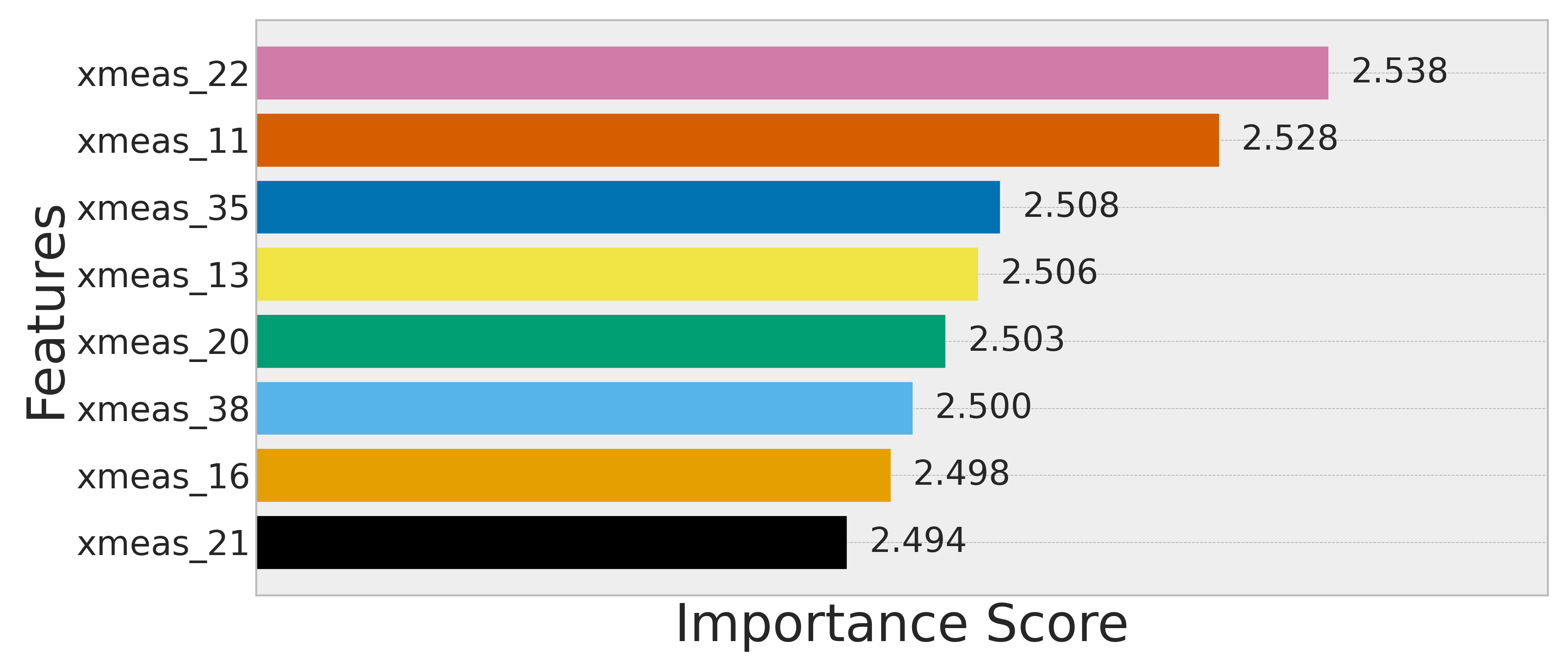}}
    \caption{This plot represents the top 8 features of the \ac{GFI} ranking
      returned by \newmethod\ on \ac{TEP}.}%
    \label{fig:Score-Plot-TEP}
\end{figure}

Figure~\ref{fig:Score-Plot-TEP} exhibits the Score Plot for the \ac{TEP}
dataset. Features \texttt{xmeas\_22} and \texttt{xmeas\_11}
emerge as more relevant than others, aligning with the ground truth~\cite{don2019dynamic}
since \texttt{xmeas\_11} is the root cause for fault \texttt{IDV12}.
Furthermore, the Sign Directed Graph in~\cite{don2019dynamic} proves also the
saliency of \texttt{xmeas\_22} (i.e.\ the Separator cooling water
outler temperature), a direct consequence of \texttt{xmeas\_11}. The causal
relationship between these two attributes leads the model to position them
on top of the \ac{GFI} ranking.

\subsubsection{Feature Selection Proxy Task}\label{sec:fs-TEP}

Utilizing labelled samples, different \ac{XAI} models are quantitatively
compared on the \ac{TEP} dataset via the Feature Selection proxy task,
described in~\cite{exiffi}. Two model-specific approaches (i.e.
\ac{DIFFI}~\cite{carletti2023interpretable} and \newmethod) are considered
alongside two model-agnostic algorithms: \ourmethod~\cite{acme_ad} and
KernelSHAP~\cite{lundberg2017unified}.

All the models were employed to explain the \EIFplus\ \ac{AD} model, except for
\ac{DIFFI}, tailored to the \ac{IF} model. As described
in~\ref{sec:experimental_setup} the proxy task requires producing a \ac{GFI}
ranking. However, \ourmethod\ and KernelSHAP algorithms can only produce
\ac{LFI} scores, so those are used for their rankings \footnote{For KernelSHAP
2\% of the dataset is used as background and \ac{SHAP} values are computed on
the 100 most anomalous points}.

\begin{figure*}[t]
    \centering
    \begin{subfigure}{0.9\linewidth}
        \centering
        \includegraphics[width=\linewidth]{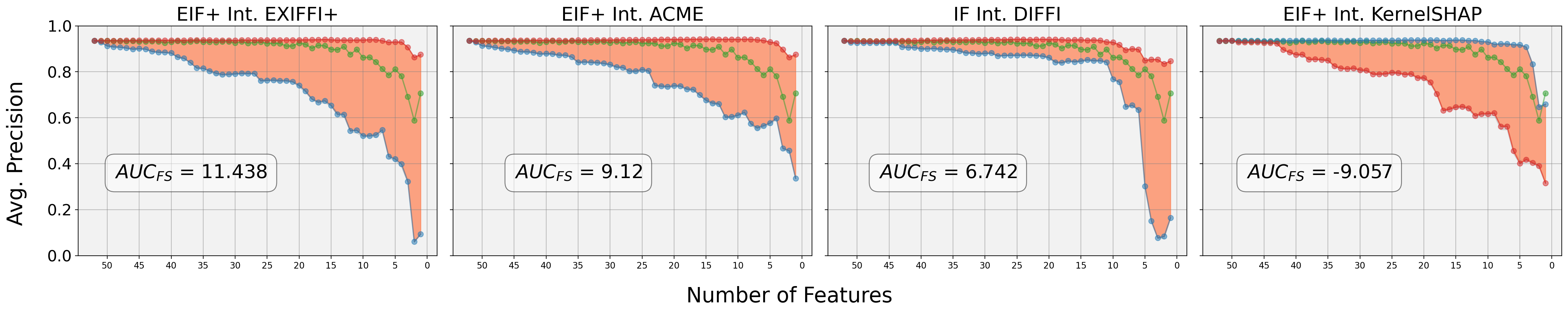}
        \caption{TEP dataset}
        \label{fig:multi_fs_plot_TEP}
    \end{subfigure}

    \vspace{0.5em}

    \begin{subfigure}{0.9\linewidth}
        \centering
        \includegraphics[width=\linewidth]{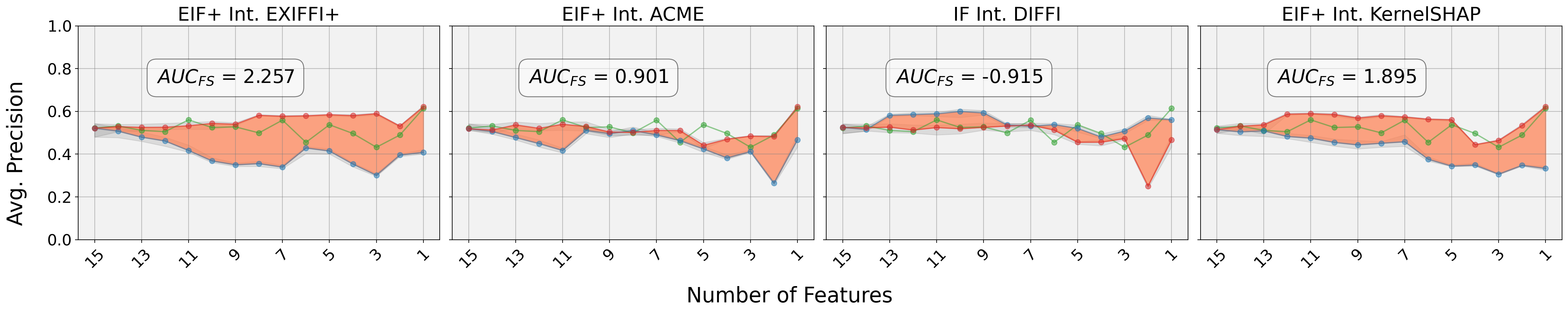}
        \caption{CoffeeData dataset}%
        \label{fig:multi_fs_plot_coffee}
    \end{subfigure}

    \caption{Feature selection results for \textit{inverse} (red), \textit{direct} (blue), and \textit{random} (green) approaches using \ac{DIFFI}, \ac{AcME-AD}, KernelSHAP, and \newmethod\ \ac{XAI}: (a) \ac{TEP}, (b) \texttt{CoffeeData}.}%
    \label{fig:multi_fs_plot}
\end{figure*}

Feature Selection plots are grouped in Figure~\ref{fig:multi_fs_plot_TEP}.
Comparing the $AUC_{FS}$ scores, the most effective explanations are produced
by \newmethod\ and \ourmethod. In these plots, the \textit{direct} approach has
a clearer decreasing trend compared to \ac{DIFFI} and KernelSHAP, resulting in
higher $AUC_{FS}$ values. KernelSHAP struggles to rank features effectively,
due to the limitations imposed by the sub-sampled background data to make it
computationally feasible, leading to a negative $AUC_{FS}$ score.


Feature Selection plots for the model-agnostic methods \ourmethod and
KernelSHAP were produced also considering the \ac{IF} model achieving similar
results \footnote{For the interested reader additional results can be consulted
at \url{https://github.com/AMCO-UniPD/exiffi-ind.git}}.

Moreover the standard deviation of the average
precision values is in the order of $10^{-3}$ confirming the statistical
significance of the results.


These results further confirm the improvement brought by \newmethod, and
consequently by \ac{EIF}-based approaches, over the other interpretability
methods considered in this study.

\subsection{Case Study II:\ \texorpdfstring{\ac{PIADE}}{PIADE} Dataset}\label{sec:results-PIADE}

For the second case study, the \ac{PIADE} dataset is considered. Differently
from \ac{TEP}, comprising synthetically generated data, \ac{PIADE} incorporates
alarm logs data coming from operating packaging machines lacking annotated
samples. Consequently, the application of the Feature Selection proxy task is
precluded, confining the experimental results to the assessment of Global
interpretability, as addressed in Sections~\ref{sec:GFI-PIADE}.%

\subsubsection{Global Feature Importance}\label{sec:GFI-PIADE}

This section reports the \ac{GFI} scores produced by \newmethod\ on \ac{PIADE}.
Only a subset of results is reported; full results are in~\cite{exiffi_rtsi}.

%
%


\begin{table}[ht]
  \centering
  \caption{Top three features according to the \ac{GFI} rankings produced in the different \ac{PIADE} machines datasets}
  \label{tab:gfi_ranking_tab}
  \resizebox{\columnwidth}{!}{
    \begin{tabular}{cccc}
      \toprule
      \textbf{Dataset} & \textbf{$1^{st}$} & \textbf{$2^{nd}$} & \textbf{$3^{rd}$} \\
      \midrule
      \texttt{piade\_s1} & \texttt{\%idle} & \texttt{A\_066} & \texttt{\%scheduled\_downtime} \\
      \texttt{piade\_s2} & \texttt{\%scheduled\_downtime} & \texttt{A\_010} & \texttt{A\_005} \\
      \texttt{piade\_s3} & \texttt{\%idle} & \texttt{\%downtime} & \texttt{A\_008} \\
      \texttt{piade\_s4} & \texttt{idle/idle} & \texttt{\%idle} & \texttt{A\_010} \\
      \texttt{piade\_s5} & \texttt{\%idle} & \texttt{idle/idle} & \texttt{A\_005} \\
      \bottomrule
    \end{tabular}
  }
\end{table}

Table~\ref{tab:gfi_ranking_tab} reports the top 3 features obtained from the
rankings induced by \ac{GFI} scores computed on the data from the 5 packaging
machines included in \ac{PIADE}. Taking  into account that these machines are
of the same type but are inserted in different working and environmental
conditions the result show an high consistency in detecting features regarding
idle and downtime intervals as important descriptors. On the other hand the
differences in the operational environment of the machines introduces some
variability in the specific alarm features included in the top 3 rankings.

\subsection{Case Study III:\ \texttt{CoffeeData} Dataset}\label{sec:results-coffe}

This section reports results on \texttt{CoffeeData}, which differs from
the other datasets due to the time-dependent nature of its data.

\subsubsection{Global Feature Importance}\label{sec:GFI-coffe}

Global interpretability on \texttt{CoffeeData} is shown via a visualization
designed to highlight the temporal structure of the data.

\begin{figure}[htbp]
    \centering
    \centerline{\includegraphics[width=0.95\linewidth]{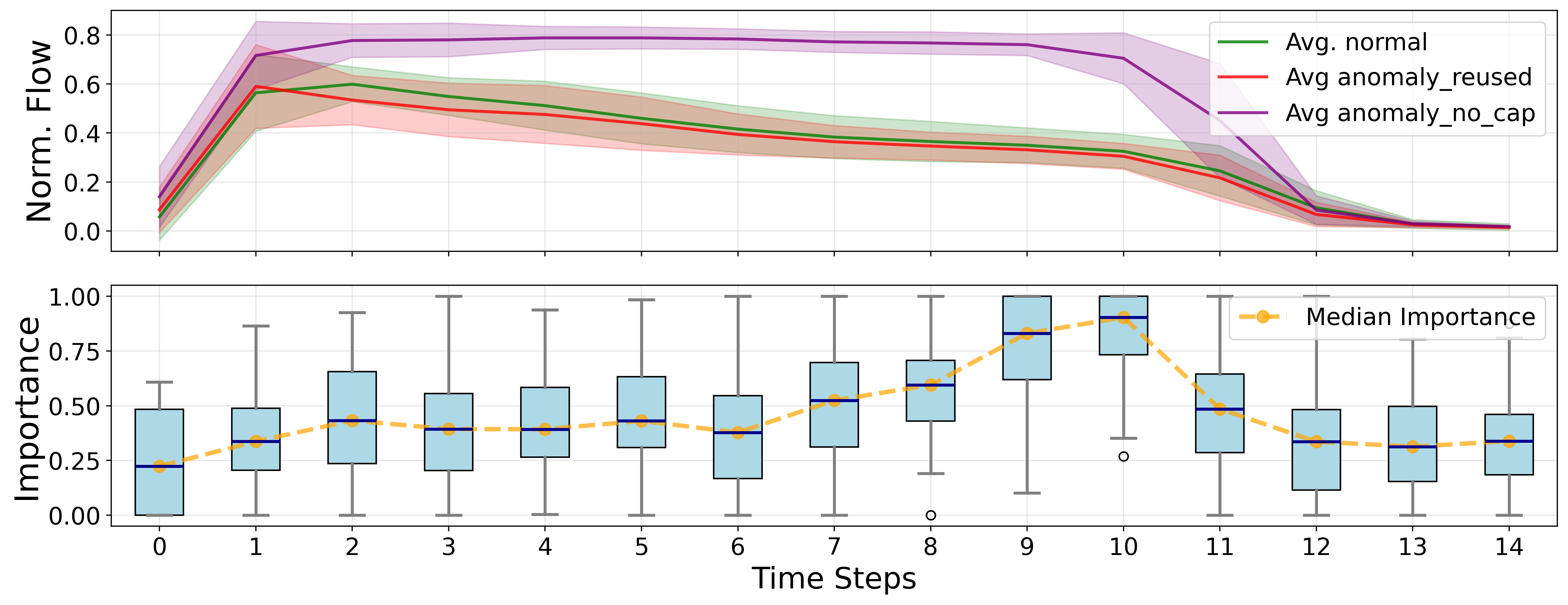}}
    \caption{The trend of the \ac{GFI} scores increases in the time interval where there is the most significant difference between the time series of normal and anomalous capsules}%
    \label{fig:time_gfi_plot}
\end{figure}

The top half of Figure~\ref{fig:time_gfi_plot} reports the average time series
divided into normal and anomalous data. In the bottom, the distribution of the
\ac{GFI} scores across multiple runs of each feature is represented with a
box-plot, with the median highlighted by an orange line, which shows the time
evolution of importance scores.

Analyzing the \ac{GFI} plot, a significant increase in the importance score can
be observed between sample 8 and sample 11, which correspond to the highest
difference between normal and anomalous time series, as observable in~\ref{fig:time_gfi_plot}. This result confirms the ability of the \newmethod\
interpretation to identify a good moment to signal the user about an
anomalous usage of the equipment.

\subsubsection{Feature Selection Proxy Task}\label{sec:fs-coffe}

Figure~\ref{fig:multi_fs_plot_coffee} depicts the Feature Selection plots for
the four interpretations under analysis. The plot is produced with the same
settings employed in~\ref{sec:fs-TEP} and confirms the superiority of
\newmethod\ in ranking features in terms of importance with respect to other
state-of-the-art approaches. Notably, \ourmethod\ and \ac{DIFFI} produce a
negative $AUC_{FS}$ score meaning that useless features were wrongly ranked in
top positions.\ Similarly to \ac{TEP} also in this case the low variation of the
average precision values over different runs proves that the results are
statistically significant.

\subsection{Time Comparison Experiment}\label{sec:timecomparison}

Time efficiency is a key requirement in the deployment of \ac{ML} models
in \ac{IIoT}. Alarms described in \ac{PIADE} may be triggered at a frequency
of four alarms per minute~\cite{FORMULA}. Consequently, \ac{AD} models should
swiftly detect anomalies to avoid catastrophic events. Accordingly, in this section
the time efficiency of \newmethod\ is compared to the one of \ac{DIFFI}, \ourmethod\ and
KernelSHAP, introduced in~\ref{sec:related_works}.

The experiment assesses the time taken by each one of the models
under examination to generate \ac{LFI} explanation for a single
anomalous point\footnote{The experiments were performed using an Intel i5
processor with 4 cores, 64 bit, 2.8 GHz, RAM 16 GB}. Due to limited
computational resources, KernelSHAP could only use sub-sampled versions of
\ac{TEP} and \ac{PIADE} (2\% and 25\% respectively) as the \textit{background}
data used to fit the explainer.

Table~\ref{tab:tab_time_comparison} outlines the time performances of the four
models considered on both datasets. Model-specific approaches (i.e.\ \ac{DIFFI}
and \newmethod) exhibit efficient computational performance, while
model-agnostic models (i.e.\ \ourmethod\ and KernelSHAP) demonstrate significantly
lower efficiency.

KernelSHAP's high computational burden makes it impractical for industrial use,
especially since it requires sub-sampling the dataset, reducing explanation accuracy.

Comparing the computational performances of the three datasets, \ac{PIADE}
exhibits higher time values: its high feature count highly affects the
KernelSHAP and \ourmethod\ complexities. \texttt{CoffeeData} is the dataset with
the lowest number of attributes and it is in fact associated to the faster
explanation times.


As analyzed in~\cite{acme_ad}, the asymptotic complexity of KernelSHAP grows
exponentially with the number of features, while \ourmethod\ exhibits linear
complexity with respect to the number of quantiles $Q$ used for input
perturbation and the feature count $p$.


The computational complexity of \ac{DIFFI} and \newmethod\ derives from the
structure of isolation trees and depends on the bootstrap sample size $\psi$,
the number of trees $T$, and the number of features $p$. For \ac{DIFFI}, which
relies on the single-axis partitions of \ac{IF}, the complexity is
$\mathcal{O}(T \cdot \psi \cdot \log(\psi))$. For \newmethod, the complexity
increases to $\mathcal{O}(T \cdot p \cdot \psi \cdot \log(\psi))$ due to the
consideration of $p$-dimensional hyperplane partitions.

Consequently, \ac{DIFFI} and \newmethod\ maintain exceptional efficiency
independently of feature count with the only drawback of being tailored
to specific models.


Across all datasets, \newmethod\ achieves approximately one order of magnitude
speedup compared to \ac{DIFFI}. This performance improvement is attributable to
the implementation of critical code segments in the C programming language
leveraging parallel computing techniques. Under ideal conditions of perfect load
balancing and negligible synchronization overhead, the asymptotic complexity of
\newmethod\ is reduced by a factor of $P$, where $P$ denotes the number of
available processing cores.




\subsection{Raspberry Pi Inference Experiments}\label{subsec:raspberry_exp}

To assess the industrial applicability of the proposed method, inference
experiments were conducted on an edge device: the Raspberry PI 3 Model B,
which reflects the hardware characteristics of devices commonly used in the
\ac{IIoT} sector. Fit, predict, and importance times are reported in
Table~\ref{tab:time_table_rasberry}.

The computational times are slightly higher than those presented in
Tables~\ref{tab:tab_time_comparison},~\ref{tab:metrics_table_tep},
and~\ref{tab:metrics_table_smd}, as those experiments were performed on more
powerful hardware. Nevertheless, the observed times remain acceptable for
industrial deployment in systems such as \ac{SCADA} and \ac{PLC}.


\newcolumntype{A}{>{\centering}p{0.09\textwidth}}
\newcolumntype{C}{>{\centering\arraybackslash}p{0.07\textwidth}}

\begin{table}[ht]
  \centering
  \caption{Inference times on Rasberry PI 3 Model B on \texttt{CofeeData}. Best results in \textbf{bold}.}
  \label{tab:time_table_rasberry}
  \begin{tabular}{cccc}
    \toprule
    \textbf{Model} &
    \textbf{Fit}        [s]    &
    \textbf{Predict}    [s]    &
    \textbf{Importance} [s]    \\
    \midrule
    \texttt{IF} & 8.63 & \textbf{0.021} & 0.24 \\
    \texttt{EIF} & \textbf{6.43} & \textbf{0.021} & \textbf{0.055} \\
    \EIFplus & 7.55 & 0.022 & \textbf{0.055} \\
    \bottomrule
  \end{tabular}
\end{table}

\begin{table}[t]
\centering
\caption{
LFI explanation times (in seconds) for a single sample by \newmethod, \ourmethod, DIFFI, and KernelSHAP on \ac{TEP}, \ac{PIADE}, and \texttt{CoffeeData} datasets.
}
\label{tab:tab_time_comparison}
\small
\begin{tabular}{cccc}
\toprule
\textbf{Method} & \textbf{\ac{TEP}} & \textbf{\ac{PIADE}} & \textbf{\texttt{CoffeeData}} \\
\midrule
\newmethod\     & \textbf{0.016}   & \textbf{0.017}   & \textbf{0.014} \\
DIFFI          & 0.082            & 0.070            & 0.107 \\
\ourmethod     & 3.60             & 16.5             & 0.523 \\
KernelSHAP     & 112.42           & 138.94           & 4.99 \\
\bottomrule
\end{tabular}
\end{table}

\subsection{Synthetic Datasets Studies}\label{sec:syn_data_exp}

\begin{figure}[htbp]
    \centering
    \includegraphics[width=0.8\linewidth]{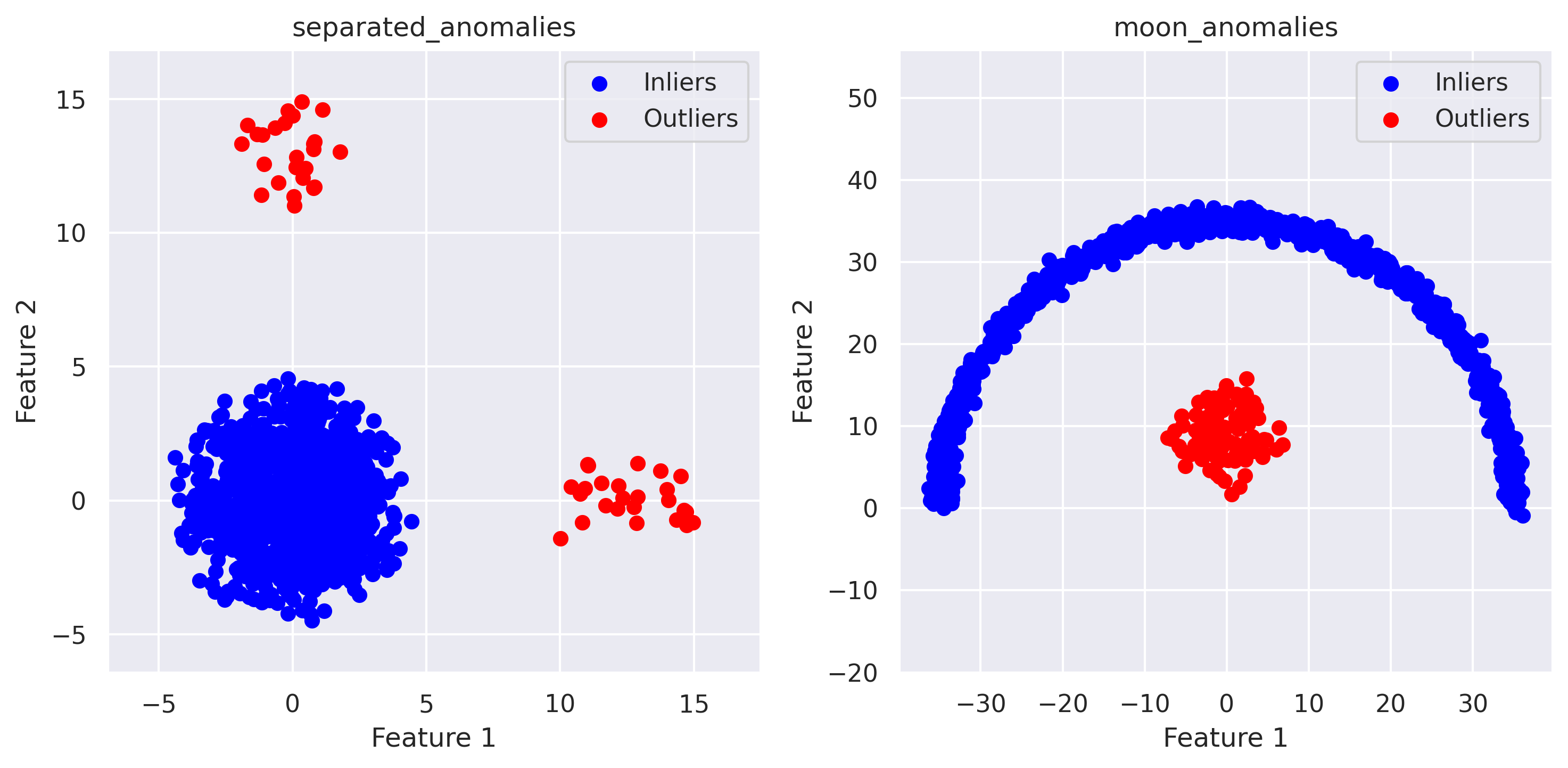}
    \caption{Scatter plots representing the synthetic datasets: \texttt{xy\_axis} on the left and \texttt{half\_moon} on the right}
    \label{fig:syn_scatter_plot}
\end{figure}

\begin{figure}[htbp]
    \centering
    \includegraphics[width=0.6\linewidth]{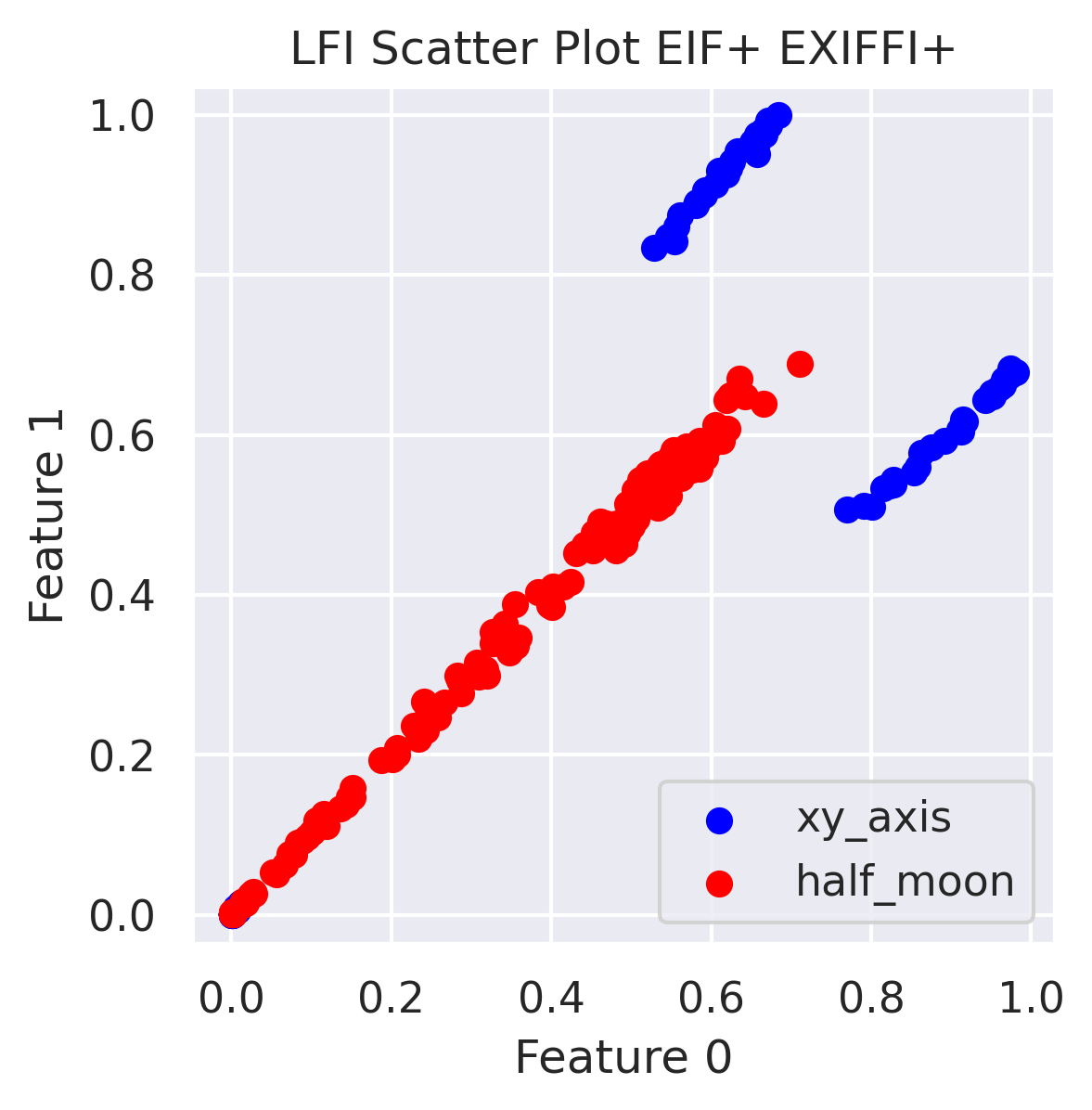}
    \caption{\ac{LFI} scatter plot for synthetic datasets: \texttt{xy\_axis} in blue and \texttt{half\_moon} in orange}
    \label{fig:lfi_scatter_plot}
\end{figure}

Two synthetic datasets are used to assess \newmethod's ability to explain
root causes of anomalies from multivariate interactions. Synthetic data
provide ground truth on anomaly-inducing features, enabling rigorous verification.
The two datasets represent distinct scenarios: the
\texttt{xy\_axis} dataset (left figure in~\ref{fig:syn_scatter_plot})
represents the case in which two anomaly clusters are aligned along two
different features, i.e., anomalies associated with two separate faults;
whereas the \texttt{half\_moon} dataset (right figure
in~\ref{fig:lfi_scatter_plot}) showcases a single cluster of anomalies enclosed
within a moon-shaped region of inliers.

The methodology employed to generate these datasets is inspired by the approach
proposed in~\cite{exiffi}.

\begin{figure}[htbp]
    \centering
    \includegraphics[width=0.8\linewidth]{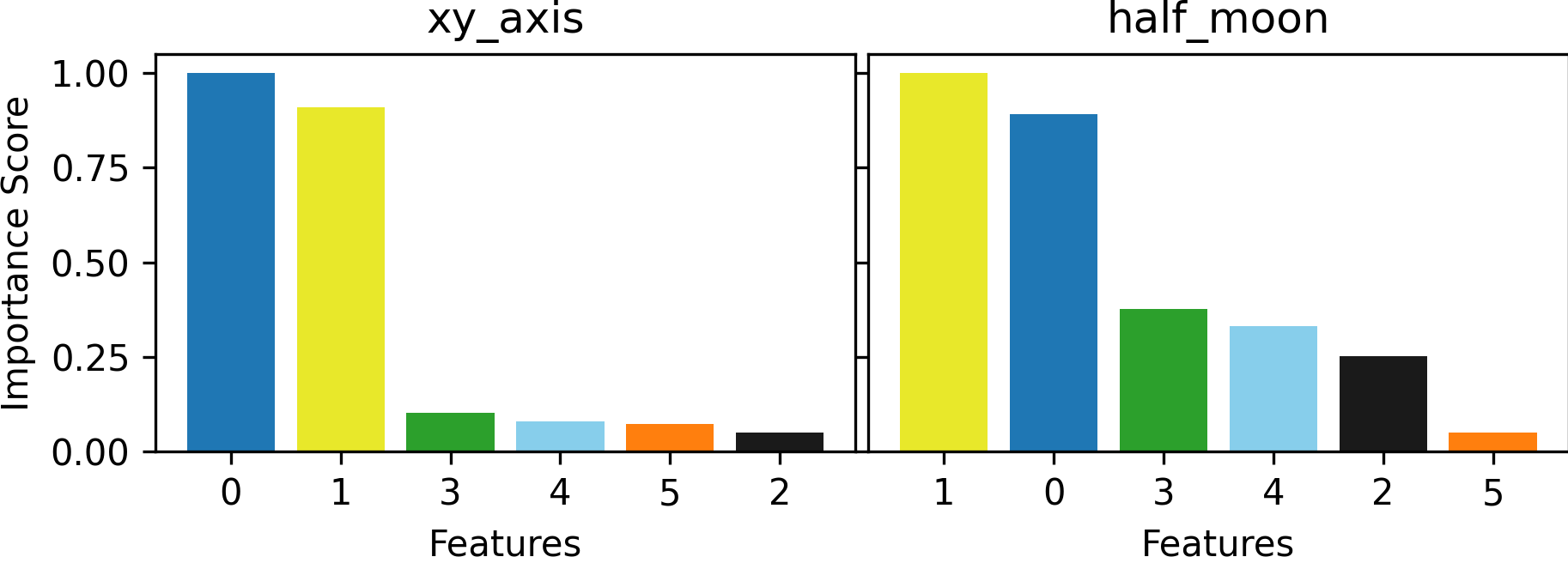}
    \caption{Normalized \ac{GFI} scores on synthetic datasets: \texttt{xy\_axis} on the left and \texttt{half\_moon} on the right.}
    \label{fig:syn_data_score_plots}
\end{figure}

The \ac{GFI} rankings obtained for \texttt{xy\_axis} and \texttt{half\_moon} are reported in Figure~\ref{fig:syn_data_score_plots}.

Figure~\ref{fig:syn_data_score_plots} shows similar rankings between the two datasets, with features 0 and 1 ranked highest.
However, when considering the Feature Selection proxy task results, described in
Section~\ref{sec:fs-TEP}, the outcomes diverge significantly.

\begin{figure}[htbp]
    \centering
    \includegraphics[width=0.9\linewidth]{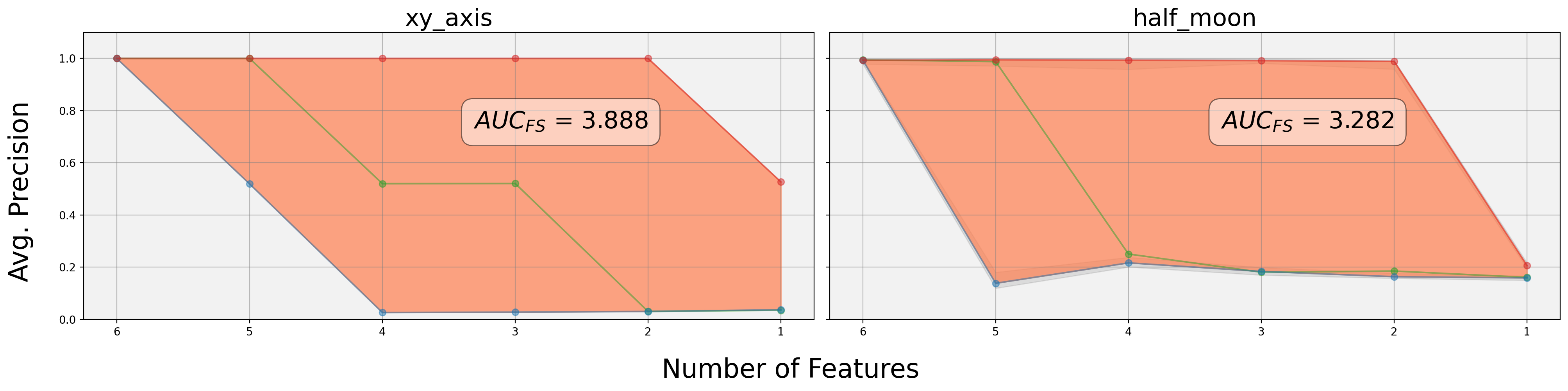}
    \caption{\ac{GFI} feature selection plots for synthetic datasets: \texttt{xy\_axis} on the left and \texttt{half\_moon} on the right}
    \label{fig:syn_data_fs_plot}
\end{figure}

Specifically, for \texttt{xy\_axis} (left figure in~\ref{fig:syn_data_fs_plot}),
the two anomaly clusters are independently caused by different features. Removing
one of them from the input space (e.g., feature 1) results in a reduction of
average precision; however, the model can still detect anomalies since the
cluster aligned along feature 0 remains identifiable.

Conversely, for \texttt{half\_moon} (right figure
in~\ref{fig:syn_data_fs_plot}), features 0 and 1 are jointly relevant.
Excluding either feature from the feature space causes the average precision to
drop to approximately zero, as anomalies cannot be detected without both
features.

The Feature Selection plots may yield misleading interpretations: for
\texttt{xy\_axis}, the equal-sized clusters cause average precision to be halved
when one key feature is removed, which would differ with imbalanced outlier groups.
The \ac{LFI} scatter plots in Figure~\ref{fig:lfi_scatter_plot} provide a
more robust evaluation. These visualizations depict
the \ac{LFI} scores for features 0 and 1 across the two scenarios.

As expected, \texttt{xy\_axis} (in blue) reveals two well-separated clusters:
one group exhibits high \ac{LFI} scores on feature 0 and low scores on feature
1, and vice versa. In contrast, \texttt{half\_moon} (in orange) shows that the
\ac{LFI} scores are aligned along the bisector line of the subspace formed by
the two important features, demonstrating their joint importance in anomaly
detection.


\subsection{Ablation Studies}\label{sec:ablation}

In this section ablation studies are performed on some key \ac{EIF} and
\newmethod\ hyperparameters to assess their influence on model performances. The
considered hyperparameters are the number of isolation trees used to fit the
ensemble and the contamination factor (i.e.\ the percentage of anomalies in the
data). The latter has a crucial importance in industrial settings where
labelled data are not available and thus the contamination factor has to be set
without any objective grounding.

Ablation studies were conducted also on other hyperparameters (i.e.
\texttt{max\_depth} and \texttt{max\_samples}) but the results demonstrate
their minimal dependence with detection performances (i.e. average precision)
and are thus not included in this study. For the interested reader, ablation studies
on the $\eta$ hyperparameter are reported in~\cite{exiffi}.

In these experiments we observed the trend of some \ac{AD} and interpretability
metrics for different hyperparameters values. For this reason these were
conducted on datasets containing ground truth knowledge on the anomalous data
like \ac{TEP} and \texttt{CoffeeData}.

For the sake of space only the results on the \ac{TEP} dataset are reported but
the ones obtained on \texttt{CoffeeData} are not much different.

\subsubsection{Number of Trees}\label{subsubsec:ablation_trees}

This experiment consists in tracking the variation of the average precision
\ac{AD} metric as the number of trees used to fit the underlying \ac{AD} model
increases. More trees expose the model to higher data variability, improving anomaly
detection. Average precision is expected to increase with the ensemble size,
at the cost of longer fitting and prediction times.

\begin{figure}[htbp]
    \centering
    \centerline{\includegraphics[width=0.7\linewidth]{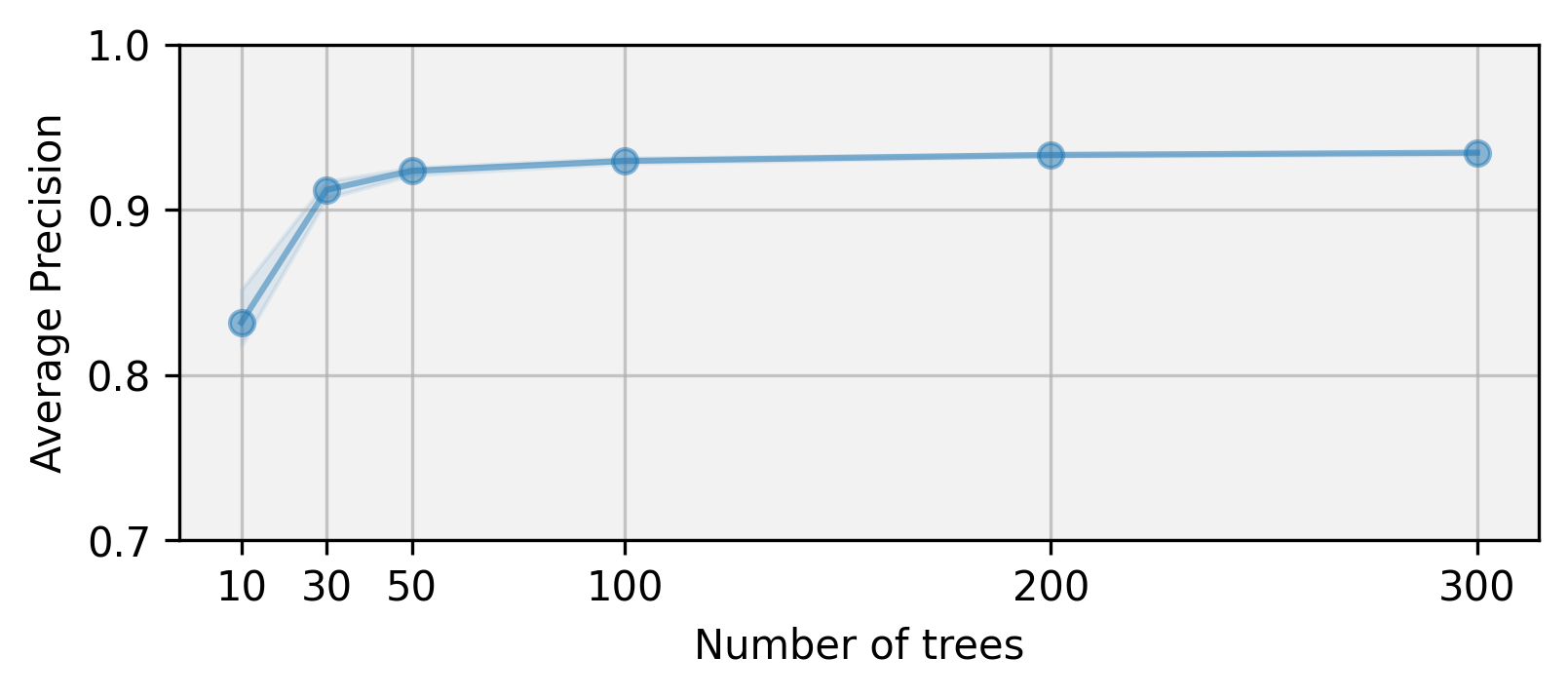}}
    \caption{The average precision metric
    increases as the number of trees used to fit the \ac{AD} model increases}%
    \label{fig:ablation_trees_plot}
\end{figure}

Figure~\ref{fig:ablation_trees_plot} shows improvement from 10 to 30 trees,
with performance saturating at~93\% average precision beyond 50 trees,
suggesting 50 trees suffice for \ac{TEP}. Computational time grows linearly
with the number of estimators (not shown for space).

\subsubsection{Contamination Prediction}\label{subsubsec:ablation_cont_prediction}

In \ac{AD} models the contamination factor is used to set a threshold on the
anomaly scores in order to convert them into binary label to classify a point
as anomalous or not. Consequently is crucial to correctly set this quantity,
especially in industrial settings.

We vary the contamination level and measure ROC AUC of \EIFplus, testing values
both above and below the true contamination ($\approx 4\%$ for \ac{TEP}),
evaluating the effect of false negatives (underestimation) and false positives
(overestimation).

\begin{figure}[htbp]
    \centering
    \centerline{\includegraphics[width=0.7\linewidth]{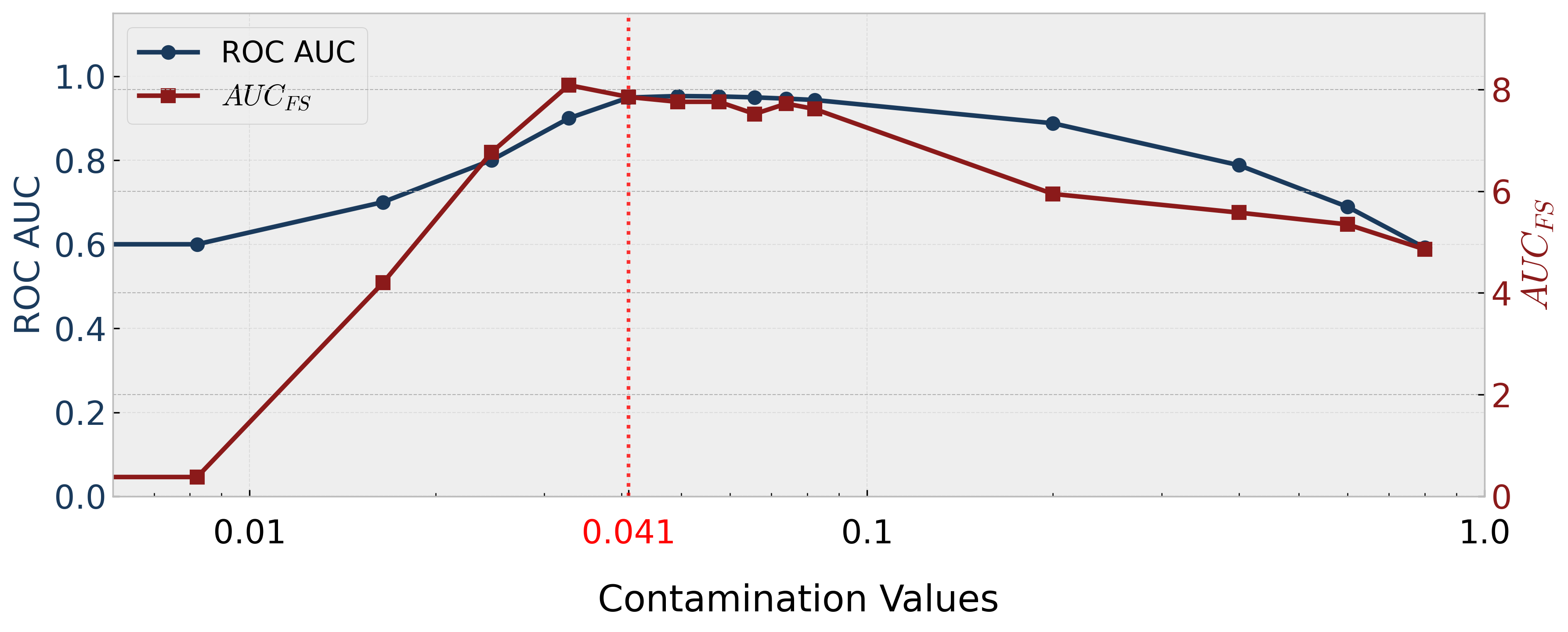}}
    \caption{Impact of contamination level on the ROC AUC score (top figure) and on the $AUC_{FS}$ metric. In both cases the best performances are obtained near the true dataset contamination value, indicated by the red vertical line. The x-axis is visualized in logarithmic scale.}%
    \label{fig:multi_ablation_cont_plot}
\end{figure}

The top plot of Figure~\ref{fig:multi_ablation_cont_plot} shows the impact of contamination level on ROC AUC performance, exhibiting a bell-shaped curve whose maximum occurs at approximately the true dataset contamination value.
Notably the curve is almost symmetric, meaning that a significant overestimation produces a similar
effect on the ROC AUC score as an underestimation.

\subsubsection{Contamination Feature Selection}\label{subsubsec:ablation_cont_fs}

This ablation evaluates interpretability via the $AUC_{FS}$ metric
(Section~\ref{sec:experimental_setup}), again varying contamination since it
affects \ac{GFI} computation (Section~\ref{sec:proposed_approach}),
using the same values as in Section~\ref{subsubsec:ablation_cont_prediction}.

The bottom plot of Figure~\ref{fig:multi_ablation_cont_plot} shows a similar
U-shaped curve, but $AUC_{FS}$ values are much lower for underestimated
contamination, likely because too few anomalies are available for \ac{GFI}
computation. This supports the preference for overestimating contamination,
as false positives are less harmful than false negatives.


\section{Limitations and Future Work} \label{sec:limitations}

Despite the results reported in~\ref{sec:metrics}, some limitations have to be
considered for a full integration of \newmethod\ into \ac{IIoT} infrastructure.


The main limitation is the lack of flexibility: \newmethod\ is tailored to
\ac{EIF} and \EIFplus, unlike \ourmethod\ and KernelSHAP which are model-agnostic.
This limitation creates a trade-off between flexibility and efficiency.
Nonetheless, as showcased in Table~\ref{tab:tab_time_comparison}, the
specificity of \newmethod\ results in fitting, inference and explanation times
that are orders of magnitude smaller than those of \ourmethod\ and, especially,
KernelSHAP. In the context of an industrial application, where execution times
are critical, we believe this efficiency gain outweighs the lack of
flexibility, making \newmethod\ more practical than the tested alternatives.


Another limitation is the inability to detect clustered anomalies, as discussed
in~\ref{sec:proposed_approach}. This can be partially mitigated by the
bootstrapping used during isolation forest training: if clusters are small,
anomalies may be isolated in different trees, improving detectability. For
large clusters, however, anomalies become ill-defined and this mitigation
fails. Semi-supervised \ac{AD} methods~\cite{extended_balif} have shown
promising results in such cases and represent a direction for future work in
industrial settings.


\section{Conclusions} \label{sec:conclusions}

We demonstrate \newmethod\ for industrial \ac{AD}, providing interpretable model
explanations through informative visualizations that integrate algorithmic outputs
with domain expert knowledge.


\newmethod\ was tested on four datasets coming from different
industrial settings, equipped with high data dimensionality, unlabeled data
points and non linear sensor interactions. The explanations
provided by \newmethod\ aligned with the ground truth on the anomalies' root
cause features. The effectiveness of the interpretations was proved against
state-of-the-art interpretability methods through the Feature Selection proxy
task, as shown in Sections \ref{sec:fs-TEP},\ref{sec:fs-coffe}.

\newmethod\ outperforms competing interpretability methods in computational
efficiency (Section~\ref{sec:timecomparison}), a key requirement in industrial
settings. This stems from being model-specific to \ac{EIF}, one of the most
time-efficient \ac{AD} models.

The computational efficiency and low memory footprint of \newmethod\ make it
suitable for deployment in a TinyML~\cite{TinyMLSurvey} \ac{IIoT} framework,
as showcased in~\ref{subsec:raspberry_exp},
positioning it as a practical real-time \ac{AD} and feature importance tool
for Industry 5.0.

\bibliographystyle{IEEEtran}
\bibliography{citations}{}

\end{document}